\documentclass[10pt, conference]{IEEEtran}
\usepackage[T1]{fontenc}
\usepackage[OT1]{fontenc}

\usepackage{listings}
\usepackage[dvipdfmx]{graphicx,color}
\usepackage{cite}
\usepackage{amsmath,amssymb,amsfonts}
\usepackage{algorithmic}
\usepackage{url}

\usepackage{array}
\newcolumntype{P}[1]{>{\centering\arraybackslash}p{#1}}

\begin{document}
%
\title{A Fast Content-Based Image Retrieval Method Using Deep Visual Features}


\author{\IEEEauthorblockN{Hiroki Tanioka}
\IEEEauthorblockA{Tokushima University, Japan\\
tanioka.hiroki@tokushima-u.ac.jp}
}


%


\maketitle

\begin{abstract}
Fast and scalable Content-Based Image Retrieval using visual features is required for document analysis, Medical image analysis, etc. in the present age. Convolutional Neural Network (CNN) activations as features achieved their outstanding performance in this area. Deep Convolutional representations using the softmax function in the output layer are also ones among visual features. However, almost all the image retrieval systems hold their index of visual features on main memory in order to high responsiveness, limiting their applicability for big data applications. In this paper, we propose a fast calculation method of cosine similarity with L2 norm indexed in advance on Elasticsearch. We evaluate our approach with ImageNet Dataset and VGG-16 pre-trained model. The evaluation results show the effectiveness and efficiency of our proposed method.
\end{abstract}


%
\IEEEpeerreviewmaketitle

\section{Introduction}

Vector quantization using machine learning method and deep features from Deep Learning were devised and there are expected to improve accuracy of multimedia information retrieval increasing, but the scalability was still considered an issue.
Specifically, we need large-scale content-based information retrieval method for some tasks of multimedia information retrieval with image queries, detecting copyright violation with imaged documents, and information retrieval task with word2vec~\cite{Word2VecWebsite} using bag-of-words on vector space model.
However, scale up is difficult in calculating cost of high dimension vector.
In addition, since search engines using inverted indices are premised on the inner product of vector space model for scoring, there are also problems with accuracy. Specifically, in the inner product scoring, there is a problem that the same contents of images and documents do not become the top of the search results.
If you want to get the same image on the top of the search result, it is sufficient to search for a perfect match by feature value extraction by hash value or vector quantization. In order to make an image with different background for the same subject the top of the search results, another high precision content-based information retrieval is desired.

One possible solution to this problem is to use memory-based approach, which indexes all image vectors in memory and calculates cosine similarity at high speed. Furthermore, in order to increase this scale, it is conceivable to make memory-based system in distributed configurations. This method requires a computer environment that can use huge and high-speed memory. This is all the more so in the case of distributed configurations.
As a method that does not require a large amount of memory, a method for dimensional compression of features is also proposed~\cite{Lux:2008:LLI:1459359.1459577, Amato:2018:LIR:3209978.3210089, LIU2017749, jimaging5030033}. In these cases, it is suggested that an expensive computer environment is not necessary though, instead, search accuracy is sacrificed.
As other methods of dimensionality reduction, including a method using transfer learning, are also proposed~\cite{7780596}. Still, the search accuracy is also encountered, and re-learning is necessary depending on the purpose of search.

Besides, a method to re-rank with top-k of search result is proposed~\cite{DBLP:conf/sigir/MuZYZY18}. And the system that can be searched realistically. However, this method is not accurate if the first search result does not contain a correct answer. In addition, there is room for improvement in terms of speed because the system needs re-ranking.
We propose a method to calculate cosine similarity, Manhattan distance, and Euclidean distance by using the inverted-index search engine.

\section{Approaches}

We once proposed an image retrieval system using VGG-16. This is one of the simplest image retrieval methods using Deep Learning~\cite{liarr2017/tanioka}.
In this system, since the inner product is used for the search score, there are cases where around the top of the search results are occupied by undesirable ones. In order to eliminate those ones, it is preferable to adopt cosine similarity as a score.

To adopt cosine similarity as a score, it is necessary to change a search engine's scoring formula. However, in an inverted-index search engine, only an inner product score can be calculated. For this reason, Cun Mu et al.~\cite{DBLP:conf/sigir/MuZYZY18} proposed a method to re-rank the search results. The accuracy deterioration is unsolved when the first search result does not include the correct answer. Also, there is a problem of the processing cost of re-ranking. although Amato et al.~\cite{Amato:2018:LIR:3209978.3210089} proposed the feature vector truncation to calculate similarity in memory. The effect of vector truncation on accuracy cannot be ignored. Besides, Liu et al.~\cite{7780596} have a strategy of making the number of dimensions overcoming the memory by dimensional compression using transfer learning, though. The problem is that it is necessary to re-learn each time for re-indexing.

Therefore, we decided to make improvements based on these problems in the following two points.

\begin{itemize}
\item{First}: high-speed score calculation
\item{Second}: truncating feature vectors
\end{itemize}

One is to reduce response time using some score calculations including cosine similarity. As a methodology for realizing high-speed similarity calculation, storing some statistics necessary to calculate score into an index in advance.
The other is the truncation of the vector without affecting the precision.

\subsection{Baseline system (Inner Product)}

The inner product (dot product) is a score expressed by the following equation, and can be processed at high speed by a search system using a inverted index based on a vector space model.
\begin{eqnarray}
  \text{dot}(\mathbf {x}, \mathbf {y}) = \sum _{i=1}^{n}x_{i}y_{i}
\end{eqnarray}
Where, $i$ is a variable for element of feature vector. $n$ is the dimension number of feature vector. $x$ means query vector and $y$ means retrieved image vector.

\begin{figure}[t]
  \centering
  \includegraphics[height=260pt]{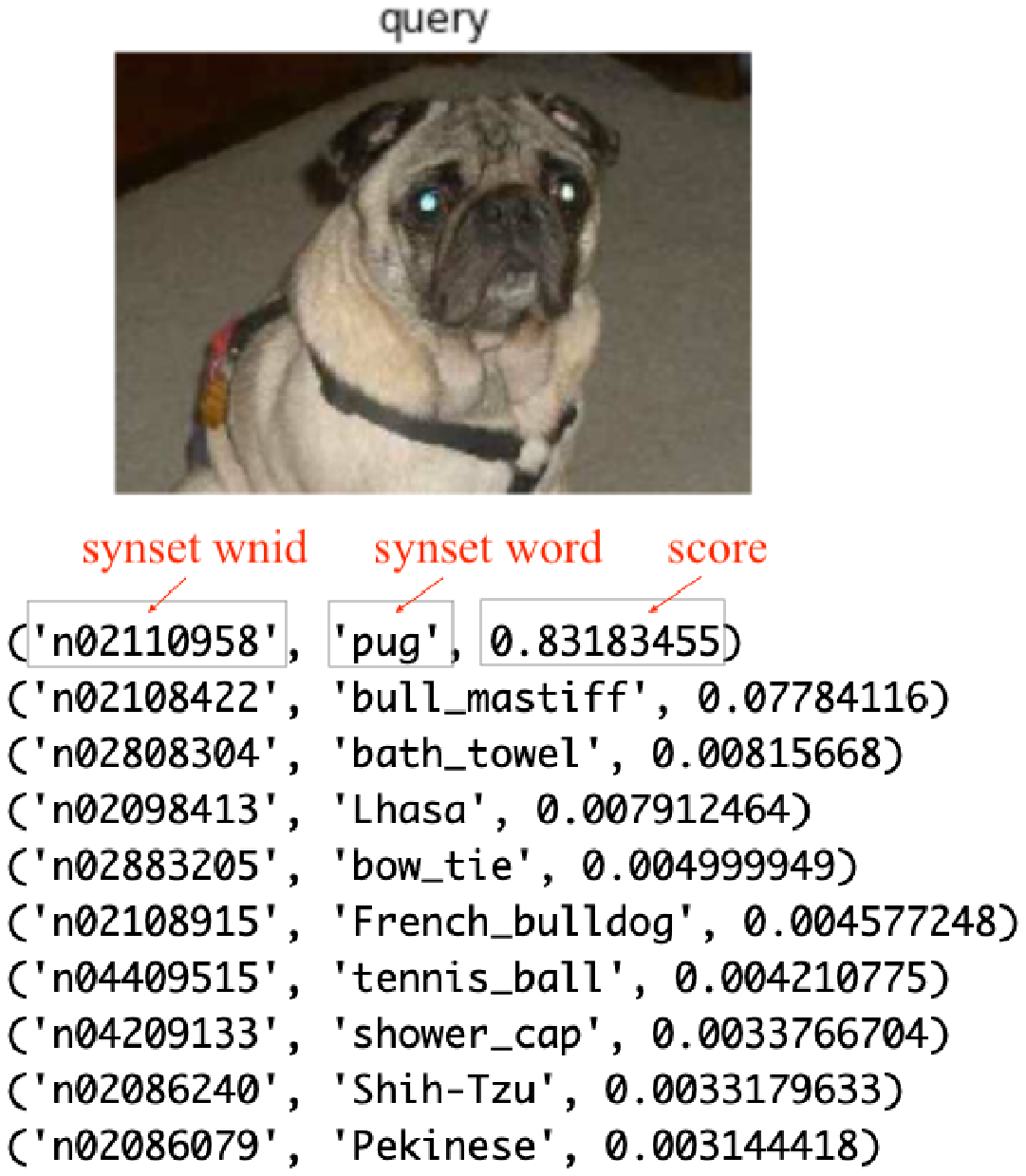}
  \caption{An image and $10$ deep visual features from VGG-16.}
  \label{fig:image}
\end{figure}

\subsection{Re-ranking system (Cosine)}

The score using cosine similarity has high affinity to the vector representation where the feature vector is output by the softmax function, though. Cosine similarity is expressed by the following equation.
\begin{eqnarray}
\text{cos}(\mathbf {x}, \mathbf {y}) & = & \frac{{\mathbf {x} \cdot \mathbf {y}}}{{ \|\mathbf {x} \|\|\mathbf {y} \|}} = \frac{\sum \limits _{i=1}^{n}{x_{i} y_{i}}}{{\sqrt {\sum \limits _{i=1}^{n}{x_{i}^{2}}}}{\sqrt {\sum \limits _{i=1}^{n}{y_{i}^{2}}}}}
\end{eqnarray}
It is difficult to calculate cosine similarity fast, because the norm of indexed image vector needs calculating sequentially. Thus, re-calculating score within top-k ($k = 10, 100, 1,000$) search result and re-ranking the search result.
\begin{table}[t]
  \small
  \centering
  \caption{Index body for each image on Elasticsearch.}
  \label{tab:index}
  \begin{tabular}{l|l|ll}
    \hline
    $f$ & \multicolumn{3}{l}{file name}  \\
    \hline
        &                 & $i$ & synset wnid \\
        &                 & $w$ & synset word \\
    $s$ & \{synset wnid\} & $s$ & score for each synset wnid \\
        &                 & $c$ & normalized score by L2 norm \\
        &                 & $ss$ & $s$ squared \\
   \hline
\end{tabular}
\end{table}
\begin{figure}[t]
  \centering
  \tiny
\begin{lstlisting}[basicstyle=\ttfamily\footnotesize, frame=single]
# synsets: synset list of query image.
# topk: top_K search result

inline = '0'
for synset in synsets:
  id = synset[0]
  score = synset[2]
  inline += "+doc['s."+id+".c'].value*"+str(score)

res = es.search(index=indexname, body={
  "size":topk,
  "query":{
    "function_score":{
      "query":{"match_all":{}},
      "script_score":{"script":{"inline":inline}}
  }}})
\end{lstlisting}
  \caption{Query using function\_score for cosine similarity on Elasticsearch.}
  \label{fig:query}
\end{figure}

\subsection{Proposed system (Cosine)}

The calculation formula of cosine similarity is divided into two phases. The $\text{L2}$ norm of the vector is calculated and registered in the index at the indexing phase, shown in Talbe~\ref{tab:index}. Then, cosine similarity is calculated at the search phase, shown in Figure~\ref{fig:query}.
\begin{eqnarray}
\text{L2}_{\mathbf{y}} & = & \|\mathbf {y} \| = \sqrt {\sum \limits _{i=1}^{n}{y_{i}^{2}}} \\
{y^{\prime}_{i}} & = & \frac{y_{i}}{\text{L2}_{\mathbf{y}}} = \frac{y_{i}}{\sqrt {\sum \limits _{i=1}^{n}{y_{i}^{2}}}}
\end{eqnarray}
Where, $y^{\prime}$ means an image feature vector normalized by $\text{L2}$ norm. This is the same as calculating a unit vector.
\begin{eqnarray}
\text{cos}(\mathbf {x}, \mathbf {y}) & = & \frac{{\mathbf {x} \cdot \mathbf {y}}}{{ \|\mathbf {x} \|\|\mathbf {y} \|}}  \propto  \frac{{\mathbf {x} \cdot \mathbf {y}}}{{ \|\mathbf {y} \|}}
 =  \frac{\sum \limits _{i=1}^{n}{x_{i} y_{i}}}{\sqrt {\sum \limits _{i=1}^{n}{y_{i}^{2}}}} \\
& = & \sum \limits _{i=1}^{n} {x_{i}} \frac{y_{i}}{\sqrt {\sum \limits _{i=1}^{n}{y_{i}^{2}}}}
 = \sum \limits _{i=1}^{n} {x_{i}} {y^{\prime}_{i}}
\end{eqnarray}

\begin{table*}[t]
  \centering
  \tiny
  \caption{The average response time [$s$] in each feature number, $100$ top-k results, and $1.0$ resolution rate.}
  \label{tab:response}
  \begin{tabular}{l|P{13pt}P{13pt}P{13pt}P{13pt}P{13pt}P{13pt}P{13pt}P{13pt}P{13pt}P{13pt}P{13pt}P{13pt}P{13pt}P{13pt}P{13pt}P{13pt}P{13pt}c}
    \hline
Feature Number & $1$ & $2$ & $3$ & $4$ & $5$ & $6$ & $7$ & $8$ & $9$ & $10$ & $20$ & $30$ & $40$ & $50$ & $100$ & $200$ & $400$ & $1000$ \\
\hline
dot product & $0.231$ & $0.325$ & $0.274$ & $0.247$ & $0.242$ & $0.361$ & $0.367$ & $0.366$ & $0.371$ & $0.421$ & $0.740$ & $0.694$ & $0.804$ & $0.860$ & $1.429$ & $2.015$ & $2.855$ & $2.865$ \\
Manhattan(L1) & $0.273$ & $0.272$ & $0.301$ & $0.314$ & $0.313$ & $0.350$ & $0.320$ & $0.296$ & $0.352$ & $0.350$ & $0.569$ & $0.797$ & $0.696$ & $0.717$ & $1.350$ & $1.576$ & $2.583$ & $2.846$ \\
Euclid(L2) & $0.260$ & $0.283$ & $0.279$ & $0.298$ & $0.289$ & $0.317$ & $0.313$ & $0.318$ & $0.331$ & $0.345$ & $0.508$ & $0.633$ & $0.798$ & $0.988$ & $1.971$ & $3.796$ & $5.778$ & $6.556$ \\
cosine & $0.226$ & $0.252$ & $0.269$ & $0.271$ & $0.284$ & $0.295$ & $0.306$ & $0.320$ & $0.331$ & $0.354$ & $0.639$ & $0.620$ & $0.614$ & $0.589$ & $1.136$ & $1.434$ & $2.597$ & $3.075$ \\
dot+cos & $44.041$ & $42.209$ & $39.723$ & $36.100$ & $34.998$ & $35.756$ & $35.390$ & $35.795$ & $35.693$ & $36.072$ & $35.446$ & $35.089$ & $37.930$ & $41.258$ & $46.971$ & $46.536$ & $51.972$ & $53.013$ \\     \hline
\end{tabular}
\end{table*}
\begin{table}[t]
  \centering
  \tiny
  \caption{Mean Average Precision (MAP) in each feature number, $100$ top-k results, and $1.0$ resolution rate.}
  \label{tab:feature}
  \begin{tabular}{l|P{6pt}P{6pt}P{6pt}P{6pt}P{6pt}P{6pt}P{6pt}P{6pt}P{6pt}P{6pt}c}
    \hline
    Feature Number & $1$ & $2$ & $3$ & $4$ & $5$ & $6$ & $7$ & $8$ & $9$ & $10$ & $20$  \\
    \hline
dot product & $0.037$ & $0.046$ & $0.046$ & $0.046$ & $0.046$ & $0.046$ & $0.046$ & $0.055$ & $0.055$ & $0.055$ & $0.054$ \\
Manhattan(L1) & $0.950$ & $1.000$ & $1.000$ & $1.000$ & $1.000$ & $1.000$ & $1.000$ & $1.000$ & $1.000$ & $1.000$ & $1.000$ \\
Euclid(L2) & $0.917$ & $1.000$ & $1.000$ & $1.000$ & $1.000$ & $1.000$ & $1.000$ & $1.000$ & $1.000$ & $1.000$ & $1.000$ \\
cosine & $0.043$ & $0.390$ & $0.759$ & $0.853$ & $0.842$ & $0.925$ & $1.000$ & $1.000$ & $1.000$ & $1.000$ & $1.000$ \\
dot+cos & $0.004$ & $0.038$ & $0.086$ & $0.086$ & $0.103$ & $0.069$ & $0.061$ & $0.061$ & $0.061$ & $0.078$ & $0.078$ \\
   \hline
\end{tabular}
\end{table}
\begin{figure}[t]
  \centering
  \includegraphics[width=\linewidth,height=140pt]{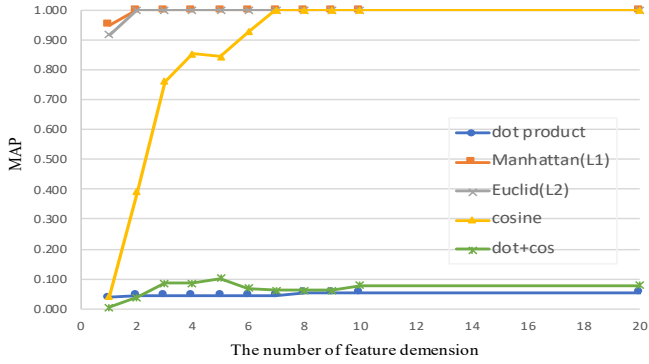}
  \caption{Comparison Mean Average Precision (MAP) of feature numbers in $100$ top-k results and $1.0$ resolution rate.}
  \label{fig:feature}
\end{figure}

\subsection{Proposed system (Manhattan)}

It is also possible to introduce Manhattan distance (City Block Distance) instead of inner product.
\begin{eqnarray}
\text{man}(\mathbf {x} ,\mathbf {y} )=\|\mathbf {x} -\mathbf {y} \|_{1}=\sum _{i=1}^{n}|x_{i}-y_{i}|
\end{eqnarray}
Where, $\|\mathbf {x} -\mathbf {y} \|_{1}$ means a Manhattan distance ($\text{L1}$ norm) from a query image feature vector $x$ to an indexed image feature vector.

\subsection{Proposed system (Euclid)}

Similarly, Euclidean distance can be calculated with inner product.
%
%
\begin{eqnarray}
\text{euc}({\mathbf{x}},{\mathbf{y}}) & = & \|{\mathbf{x}}-{\mathbf{y}}\| = {\sqrt{\sum_{{i=1}}^{n}(x_{i}-y_{i})^{2}}} \\
& = & {\sqrt{\sum_{{i=1}}^{n}(x_{i}^{2}-2x_{i}y_{i}+y_{i}^{2})}} \\
& \propto & \text{L2}_{\mathbf{y}}^{2} - 2\sum_{i=1}^{n} x_{i}y_{i}
\end{eqnarray}
Where, $\|{\mathbf{x}}-{\mathbf{y}}\|$ means an Euclidean distance ($\text{L2}$ norm) from a query image feature vector $x$ to an indexed image feature vector. For search score, Manhattan distance and Euclidean distance are converted to complement.
{\bfseries The prototype system implementation is naive using script\_score of Elasticsearch (\url{https://www.elastic.co}) on MacOS 10.14.5, 2.5GHz, 16GB DDR3, 500GB SSD.}

\begin{table}[t]
  \centering
  \tiny
  \caption{Mean Average Precision (MAP) in each resolution rate, $1,000$ feature number, and $100$ top-k results.}
  \label{tab:resolution}
  \begin{tabular}{l|ccccc}
    \hline
resolution & $1$ & $0.8$ & $0.6$ & $0.4$ & $0.2$ \\
\hline
dot product & $0.052$ & $0.035$ & $0.025$ & $0.032$ & $0.012$ \\
Manhattan(L1) & $1.000$ & $0.858$ & $0.837$ & $0.310$ & $0.002$ \\
Euclid(L2) & $0.911$ & $0.555$ & $0.553$ & $0.234$ & $0.002$ \\
cosine & $1.000$ & $0.642$ & $0.662$ & $0.240$ & $0.009$ \\
dot+cos & $0.078$ & $0.008$ & $0.044$ & $0.021$ & $0.009$ \\
    \hline
\end{tabular}
\end{table}
\begin{figure}[t]
  \centering
  \includegraphics[width=\linewidth,height=140pt]{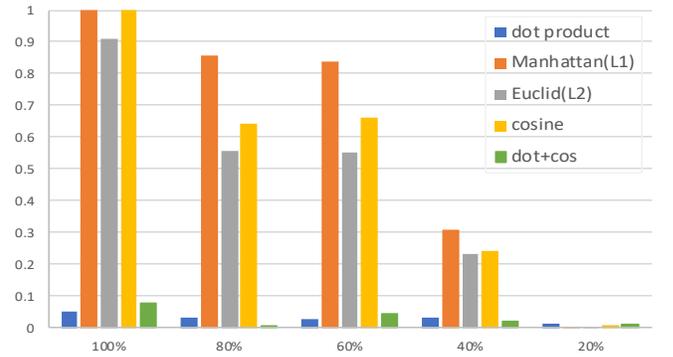}
  \caption{Comparison Mean Average Precision (MAP) of resolution rates, in $1,000$ feature number and $100$ top-k results.}
  \label{fig:resolution}
\end{figure}

\begin{figure*}[t]
  \centering
  \includegraphics[width=490pt]{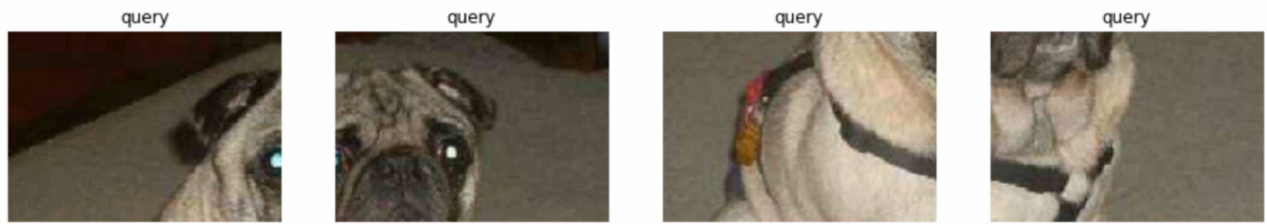}
  \vspace{-5mm}
  \caption{Quartered partial image queries for partial image search.}
  \label{fig:partial_query}
\end{figure*}

\begin{table}[t]
  \centering
  \tiny
  \caption{Mean Average Precision (MAP) in each feature number using quartered partial image query, $1,000$ feature number, and $100$ top-k results.}
  \label{tab:feature2}
  \begin{tabular}{l|P{7pt}P{7pt}P{7pt}P{7pt}P{7pt}P{7pt}P{7pt}P{7pt}P{7pt}c}
    \hline
Feature Number & $1$ & $2$ & $3$ & $4$ & $5$ & $6$ & $7$ & $8$ & $9$ & $10$ \\
\hline
dot product & $0.005$ & $0.005$ & $0.005$ & $0.004$ & $0.005$ & $0.005$ & $0.005$ & $0.005$ & $0.005$ & $0.005$ \\
Manhattan(L1) & $0.008$ & $0.008$ & $0.015$ & $0.016$ & $0.018$ & $0.020$ & $0.020$ & $0.013$ & $0.010$ & $0.011$ \\
Euclid(L2) & $0.008$ & $0.020$ & $0.020$ & $0.021$ & $0.021$ & $0.025$ & $0.025$ & $0.013$ & $0.010$ & $0.011$ \\
cosine & $0.005$ & $0.026$ & $0.051$ & $0.052$ & $0.052$ & $0.052$ & $0.052$ & $0.052$ & $0.052$ & $0.052$ \\
dot+cos & $0.001$ & $0.002$ & $0.013$ & $0.026$ & $0.026$ & $0.026$ & $0.026$ & $0.026$ & $0.026$ & $0.026$ \\
\end{tabular}
\end{table}
\begin{figure}[t]
  \centering
  \includegraphics[width=\linewidth,height=140pt]{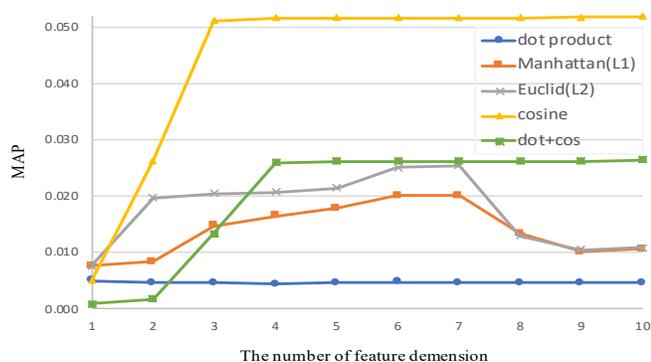}
  \caption{Comparison Mean Average Precision (MAP) of feature numbers using quartered partial image query, in $1,000$ feature number and $100$ top-k results.}
  \label{fig:feature2}
\end{figure}

\section{Experiments}

In order to evaluate the effectiveness of the proposed system, using Dog and Cat's image file~\cite{dogs-vs-cats}, We compare the image retrieval system using a baseline system using dot product score, Manhattan distance score, Euclidean distance score, cosine score, and re-ranking cosine score for its search accuracy (Mean Average Precision) and response time.

When indexing the search engine database, the recognition results of VGG-16 were used for the index. The feature quantity is obtained as a random variable having a total of $1$ in $1,000$ dimensions~\cite{ILSVRC}, shown in Figure~\ref{fig:image}. Each image is registered by adding the information, shown in Table~\ref{tab:index}. Also, when generating a search query, use information represented by Figure~\ref{fig:query}.

For the experiment, $10$ images for test query, resolution rate is $5$ levels, and quartered partial images (Figure~\ref{fig:partial_query}) are prepared (total: $250$ images). The result of response time is shown in Table~\ref{tab:response}. A result of Mean Average Precision (MAP) for each feature number is shown in Table~\ref{tab:feature} and Figure~\ref{fig:feature}. A result of Mean Average Precision (MAP) for each resolution rate is shown in Table~\ref{tab:resolution} and Figure~\ref{fig:resolution}. A result of Mean Average Precision (MAP) for each feature number using quartered partial image query is shown in Table~\ref{tab:feature2} and Figure~\ref{fig:feature2}.







\section{Conclusion}

All results show that dot product and dot+cos are behind other scores in accuracy and response time.
Figure~\ref{fig:feature} describes a fact in the task of searching for the same image, the smaller the number of dimensions, the higher the accuracy when using the Manhattan distance and the Euclidean distance. Although cosine has the same degree of precision when the number of dimensions is $7$ or more, the inner product still has low precision. We can improve it a bit with re-ranking.
Then, Figure~\ref{fig:resolution} describes that the accuracies of dot product and dot+cos are extremely low. Also, Figure~\ref{fig:feature2} describes a fact in the task of partial image search, cosine indicates the highest accuracy regardless of the number of dimensions.
The Manhattan distance and the Euclid distance are unexpectedly good at the small number of dimensions. This means cosine similarity focuses on the common points only at the angle of the vectors, while the these distances can consider the difference in vector length.
Even when targeting high-dimensional query vectors, it is possible to reduce the response time without sacrificing accuracy by adopting a top-k method that truncates vectors with small impact using cosine similarity.

In this research, it was shown that high-speed and high-precision content-based information retrieval method can be performed while suppressing the memory usage, using cosine similarity on inverted index search engine.
From now on, we want to work on machine learning algorithm using distance such as k-NN, application to clustering method such as k-means, practical application of search system using deep feature vector, and transfer learning for vector quantization suitable for the content-based information retrieval system.







\bibliographystyle{IEEEtranS}
\bibliography{icdar_wml2019}

\begin{thebibliography}{10}
\providecommand{\url}[1]{#1}
\csname url@samestyle\endcsname
\providecommand{\newblock}{\relax}
\providecommand{\bibinfo}[2]{#2}
\providecommand{\BIBentrySTDinterwordspacing}{\spaceskip=0pt\relax}
\providecommand{\BIBentryALTinterwordstretchfactor}{4}
\providecommand{\BIBentryALTinterwordspacing}{\spaceskip=\fontdimen2\font plus
\BIBentryALTinterwordstretchfactor\fontdimen3\font minus
  \fontdimen4\font\relax}
\providecommand{\BIBforeignlanguage}[2]{{%
\expandafter\ifx\csname l@#1\endcsname\relax
\typeout{** WARNING: IEEEtranS.bst: No hyphenation pattern has been}%
\typeout{** loaded for the language `#1'. Using the pattern for}%
\typeout{** the default language instead.}%
\else
\language=\csname l@#1\endcsname
\fi
#2}}
\providecommand{\BIBdecl}{\relax}
\BIBdecl

\bibitem{Amato:2018:LIR:3209978.3210089}
\BIBentryALTinterwordspacing
G.~Amato, P.~Bolettieri, F.~Carrara, F.~Falchi, and C.~Gennaro, ``Large-scale
  image retrieval with elasticsearch,'' in \emph{The 41st International ACM
  SIGIR Conference on Research \&\#38; Development in Information Retrieval},
  ser. SIGIR '18.\hskip 1em plus 0.5em minus 0.4em\relax New York, NY, USA:
  ACM, 2018, pp. 925--928. [Online]. Available:
  \url{http://doi.acm.org/10.1145/3209978.3210089}
\BIBentrySTDinterwordspacing

\bibitem{ILSVRC}
\BIBentryALTinterwordspacing
{ImageNet}. (2014) Large scale visual recognition challenge (ilsvrc). [Online].
  Available: \url{http://www.image-net.org/challenges/LSVRC/}
\BIBentrySTDinterwordspacing

\bibitem{dogs-vs-cats}
\BIBentryALTinterwordspacing
{Kaggle}. (2014) Dogs vs. cats. [Online]. Available:
  \url{https://www.kaggle.com/c/dogs-vs-cats}
\BIBentrySTDinterwordspacing

\bibitem{7780596}
H.~{Liu}, R.~{Wang}, S.~{Shan}, and X.~{Chen}, ``Deep supervised hashing for
  fast image retrieval,'' in \emph{2016 IEEE Conference on Computer Vision and
  Pattern Recognition (CVPR)}, June 2016, pp. 2064--2072.

\bibitem{LIU2017749}
\BIBentryALTinterwordspacing
H.~Liu, B.~Li, X.~Lv, and Y.~Huang, ``Image retrieval using fused deep
  convolutional features,'' \emph{Procedia Computer Science}, vol. 107, pp. 749
  -- 754, 2017, advances in Information and Communication Technology:
  Proceedings of 7th International Congress of Information and Communication
  Technology (ICICT2017). [Online]. Available:
  \url{http://www.sciencedirect.com/science/article/pii/S1877050917304349}
\BIBentrySTDinterwordspacing

\bibitem{Lux:2008:LLI:1459359.1459577}
\BIBentryALTinterwordspacing
M.~Lux and S.~A. Chatzichristofis, ``Lire: Lucene image retrieval: An
  extensible java cbir library,'' in \emph{Proceedings of the 16th ACM
  International Conference on Multimedia}, ser. MM '08.\hskip 1em plus 0.5em
  minus 0.4em\relax New York, NY, USA: ACM, 2008, pp. 1085--1088. [Online].
  Available: \url{http://doi.acm.org/10.1145/1459359.1459577}
\BIBentrySTDinterwordspacing

\bibitem{DBLP:conf/sigir/MuZYZY18}
\BIBentryALTinterwordspacing
C.~Mu, J.~Zhao, G.~Yang, J.~Zhang, and Z.~Yan, ``Towards practical visual
  search engine within elasticsearch,'' in \emph{The {SIGIR} 2018 Workshop On
  eCommerce co-located with the 41st International {ACM} {SIGIR} Conference on
  Research and Development in Information Retrieval {(SIGIR} 2018), Ann Arbor,
  Michigan, USA, July 12, 2018.}, ser. {CEUR} Workshop Proceedings,
  J.~Degenhardt, G.~D. Fabbrizio, S.~Kallumadi, M.~Kumar, A.~Trotman, Y.~Lin,
  and H.~Zhao, Eds., vol. 2319.\hskip 1em plus 0.5em minus 0.4em\relax
  CEUR-WS.org, 2018. [Online]. Available:
  \url{http://ceur-ws.org/Vol-2319/paper7.pdf}
\BIBentrySTDinterwordspacing

\bibitem{jimaging5030033}
\BIBentryALTinterwordspacing
P.~Sadeghi-Tehran, P.~Angelov, N.~Virlet, and M.~J. Hawkesford, ``Scalable
  database indexing and fast image retrieval based on deep learning and
  hierarchically nested structure applied to remote sensing and plant
  biology,'' \emph{Journal of Imaging}, vol.~5, no.~3, 2019. [Online].
  Available: \url{https://www.mdpi.com/2313-433X/5/3/33}
\BIBentrySTDinterwordspacing

\bibitem{liarr2017/tanioka}
\BIBentryALTinterwordspacing
H.~Tanioka, ``Super easy way of building image search with keras,'' Aug. 2017.
  [Online]. Available: \url{https://github.com/taniokah/liarr2017}
\BIBentrySTDinterwordspacing

\bibitem{Word2VecWebsite}
``{word2vec : Tool for computing continuous distributed representations of
  words},'' https://code.google.com/p/word2vec, [Online; accessed
  11-September-2018].

\end{thebibliography}

\end{document}